\documentclass{article}

 \usepackage[preprint]{neurips_2025}

\usepackage[utf8]{inputenc} % allow utf-8 input
\usepackage[T1]{fontenc}    % use 8-bit T1 fonts
\usepackage{hyperref}       % hyperlinks
\usepackage{url}            % simple URL typesetting
\usepackage{booktabs}       % professional-quality tables
\usepackage{amsfonts}       % blackboard math symbols
\usepackage{nicefrac}       % compact symbols for 1/2, etc.
\usepackage{microtype}      % microtypography
\usepackage{bm}
\usepackage[table]{xcolor}
\usepackage{pifont}
\usepackage{microtype}
\usepackage{graphicx}
\usepackage{subfigure}
\usepackage{booktabs} % for professional tables

\usepackage{amsmath}
\usepackage{amssymb}
\usepackage{mathtools}
\usepackage{amsthm}
\usepackage[capitalize,noabbrev]{cleveref}
\newcommand{\rbt}[1]{\textcolor{black}{#1}}
\title{RanDeS: Randomized Delta Superposition for Multi-Model Compression}

\author{Hangyu Zhou, Aaron Gokaslan, Volodymyr Kuleshov, Bharath Hariharan  \\
Computer Science, Cornell University \\
\texttt{\{hz477,akg87,vk379,bh497\}@cornell.edu} \\
\texttt{https://github.com/Zhou-Hangyu/randes}
}

\begin{document}

\maketitle

\begin{abstract}
From a multi-model compression perspective, model merging enables memory-efficient serving of multiple models fine-tuned from the same base, but suffers from degraded performance due to 
interference among their task-specific parameter adjustments (i.e., deltas).
In this paper, we reformulate model merging as a compress-and-retrieve scheme, revealing that the task interference arises from the summation of irrelevant deltas during model retrieval.
To address this issue, we use random orthogonal transformations to decorrelate these vectors into self-cancellation.
We show that this approach drastically reduces interference, improving performance across both vision and language tasks.
Since these transformations are fully defined by random seeds, adding new models requires no extra memory. Further, their data- and model-agnostic nature enables easy addition or removal of models with minimal compute overhead, supporting efficient and flexible multi-model serving.
\end{abstract}

\section{Introduction}
% \hz{The task is known. make it clear.}
Contemporary advances in machine learning are fueled by ever larger models trained on massive datasets~\citep{villalobos2022machine}. This leads to a blossom of fine-tuned models specialized for individual tasks and users' needs.
As the number of fine-tuned models proliferates, the memory cost of serving them becomes substantial. 
Compressing models with off-the-shelf tools provides a straight-forward solution, but fails to leverage the parameter redundancy between fine-tuned models with the same base~\citep{anwar2017structured,liao2023can,gholami2022survey,gou2021knowledge,yu2017compressing}. 
Delta compression alleviates this issue by decomposing fine-tuned model parameters into their pre-trained components and \emph{deltas} (parameter differences between the fine-tuned model and their base models), and only compressing individual deltas~\citep{liu2024bitdelta,ping2024delta}. But the redundancy between deltas is still left unexplored.
At the other extreme, model merging treats all task-specific adjustments as redundant and proposes to reduce all models down to a single set of weights~\citep{wortsman2022model,ilharco2022editing}. But various studies~\citep{yadav2024ties,wang2024localizing,ortiz2024task} have shown that the interference between different models, which grows more prominent with more merged models, leads to significant performance drops.

In this paper, we take a renewed look at model merging and find that for a specific task, the cause of interference are the merged deltas for the other tasks. Armed with this insight, we propose a simple framework for interference reduction with no training and minimal memory overhead.
We dub our framework \textit{Randomized Delta Superposition (RanDeS)}, 
as it works by superposing deltas with conflicting deltas randomly decorrelated into self-cancellation. Thanks to the high dimensionality of modern neural networks, we can induce orthogonality among deltas using random transformations without training. Being random also allows for specifying highly-complex transformations with memory-efficient random seeds. These features make \emph{RanDeS} preferable for multi-model serving settings with resource restrictions and dynamic user requests.

We substantiate \emph{RanDeS} with two efficient layer-wise implementations. Our first approach is to \emph{shuffle} each delta's layers before combining them, with an inverse shuffling applied at test time. Our second approach is to apply a \emph{random column-wise sign flip} or \emph{reflection} to each layer of the deltas before merging them, again inverting the transformation at test time. We test these approaches on three multi-model serving benchmarks with different use cases: up to 20 CLIP-ViT-B/32 and ViT-L/14 models for zero-shot image classification~\citep{radford2021learning}, 8 Flan-T5-base models (along with LoRA~\citep{hu2021lora} variants) for text generation~\citep{longpre2023flan}, and
7 GPT-2 for text classification~\citep{radford2019language}.
We find that across all of these benchmarks, our approach substantially improves in terms of accuracy over prior model merging-based approaches.
When compared to the original fine-tuned models, in two of the three benchmarks our approach yields near-identical accuracy to the individual models while reducing the storage costs by $4\times$.
% In sum, our contributions are:
% \hz{TBD}
% \begin{enumerate}
% \item We provide an analysis of the interference between tasks in task arithmetic, which suggests that similarity between the task vectors may be a problem.
% \item We propose two complementary strategies for reducing interference. Our first strategy randomly shuffles parameter matrices across layers. Our second strategy applies a random rotation or a sign flip to the task vectors before merging.
% \item We demonstrate through experiments on three benchmarks that our approach compresses multiple models together and achieves much higher accuracy than prior model merging based approaches.
% \end{enumerate}
\section{Problem setup}
We are given $T$ models $\{\bm{\Theta}_i\in\mathbb{R}^d\}_{i=1}^T$ fine-tuned from a pre-trained model $\bm{\Theta}_0\in\mathbb{R}^d$ on tasks $i=1, \dots, T$.
Our goal is to compress $\{\bm{\Theta}_i\}_{i=1}^T$ into a compact representation $\bm{\Theta}_* = \mathtt{compress}(\{\bm{\Theta}_i\}_{i=1}^T)$ with minimal memory usage, so that at test time, given a task $i$, we can retrieve an \emph{approximate} model $\hat{\bm{\Theta}}_i = \mathtt{retrieve}(\bm{\Theta}_*, i)$ for the task $i$ that achieves high accuracy. 

% We are given $T$ models $\{\bm{\Theta}_i\in\mathbb{R}^d\}_{i=1}^T$ fine-tuned from a pre-trained model $\bm{\Theta}_0\in\mathbb{R}^d$ on tasks $i=1, \dots, T$. Define weight delta $\bm{\Delta}_i = \bm{\Theta}_i -\bm{\Theta}_0$ (also known as \emph{task vector}).

\section{Interference in Model Merging}
Model merging approaches the problem by merging $T$ models into a single model with performance degradation due to task interference~\citep{wortsman2022model,ilharco2022editing,yadav2024ties,wang2024localizing,ortiz2024task}. Here we use \emph{task arithmetic}~\citep{ilharco2022editing}, a popular model merging framework, as an example to show what is interference and how to reduce it. The same analysis applies to other common frameworks.

Task arithmetic starts with decomposing each model $\bm{\Theta}_i$ into its pre-trained model, $\bm{\Theta}_0$, and the weight delta $\bm{\Delta}_i = \bm{\Theta}_i -\bm{\Theta}_0$ (also known as \emph{task vector}). Then individual deltas are aggregated before merging back to the shared pre-trained model weights:
\begin{equation}\label{ta_def}
    \bm{\Theta}_\star^{TA} \leftarrow \bm{\Theta}_0 + \lambda \sum_{i=1}^T \bm{\Delta}_i,
\end{equation}
where  $\lambda \in \mathbb{R}^+$ is the scaling coefficient. 
At test time, this compressed model is directly applied no matter what the task:
\begin{align} \label{ta_retrieve}
\hat{\bm{\Theta}}_i^{TA} &\leftarrow \bm{\Theta}_\star^{TA}, \nonumber\\
=&\bm{\Theta}_0 + \lambda\sum_{i=1}^T(\bm{\Theta}_i-\bm{\Theta}_0),  \nonumber\\
=&(1-\lambda)\bm{\Theta}_0 + \lambda \bm{\Theta}_i + \lambda\sum_{j\neq i}\bm{\Delta}_j.
\end{align}
The last line suggests the model being applied to task $i$ is interpolating between the pre-trained model $\bm{\Theta}_0$ and the fine-tuned model $\bm{\Theta}_i$, but with interference coming from other merged deltas: $\lambda\sum_{j\neq i}\bm{\Delta}_j$. 
To retrieve $\bm{\Theta}_i$, the first two terms suggest that we should set $\lambda$ to 1.
However this will inevitably \emph{increase} the interference.
Some prior work has tried to achieve a good balance by optimizing $\lambda$ for each model and layer using test-time adaptation~\citep{yang2023adamerging}, but performance drop still persists.

Instead of juggling with $\lambda$, we directly suppressing interference by reducing its $l_2$ norm:
\begin{equation} \label{interference_frob_norm_equation}
 \left\|\lambda \sum_{j\neq i}\bm{\Delta}_j\right\|=\lambda\sqrt{\sum_{j\neq i}\|\bm{\Delta}_i\|_2^2+2\sum_{\substack{1 \leq l < j \leq n \\ l, j \neq i}}\|\bm{\Delta}_l\|_2\|\bm{\Delta}_j\|_2\cos(\bm{\Delta}_l,\bm{\Delta}_j)}.
\end{equation}
We observe that the magnitude of interference is negatively correlated with the cosine similarity between deltas. This means we can reduce interference by decorrelating the conflicting deltas $\{\bm{\Delta}_j\}_{j\neq i}$ without altering $\bm{\Delta}_i$.
Below, we propose \emph{Randomized Delta Superposition (RanDeS)} as a principled framework for doing this.

\section{Randomized Delta Superposition (RanDeS)}\label{sec:randes}
% as it works by decorrelating the deltas with random orthogonal transformations, before merging them into a superposition. Each superposed delta is then retrieved on-call with inverse transformations while preserving the orthogonality among the conflicting deltas.
At a high level, \emph{RanDeS} first randomizes the deltas with random orthogonal transformations before merging, which decorrelates the deltas due to high dimensionality. At test time, each delta can be retrieved with its inverse orthogonal transformation while preserving the pairwise orthogonality among conflicting deltas.

\begin{figure}[h]
    \centering
    % \vspace{-4mm}
    \includegraphics[width=0.5\textwidth]{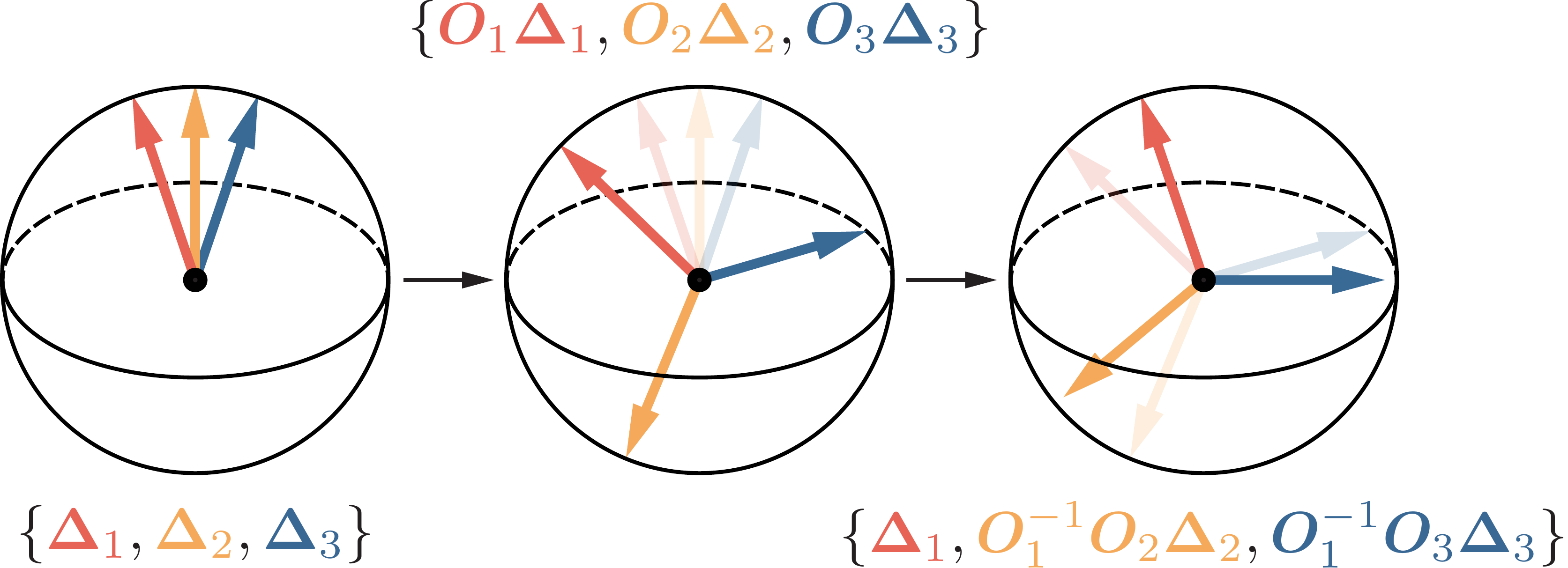}
    % \vspace{2mm}
    \caption{Illustration of \emph{RanDeS} with three deltas $\{\bm{\Delta}_i\}_{i=1}^3$.}
    \label{fig:randes_overview}
    % \vspace{-1mm}
\end{figure}

Concretely, given $T$ fine-tuned models $\{\bm{\Theta}_i\in\mathbb{R}^d\}_{i=1}^T$ and the pre-trained model $\bm{\Theta}_0\in\mathbb{R}^d$, we first derive the deltas $\{\bm{\Delta}_i\in\mathbb{R}^d\}_{i=1}^T$. Then we transform each delta $\bm{\Delta}_i$ with an orthogonal transformation $\bm{O}_i$ sampled from the orthogonal group $\mathbb{O}(d)$ before merging back to the shared pre-trained model weights:
\begin{align}
    &\bm{\Theta}_\star^{RanDeS} \leftarrow \bm{\Theta}_0 + \lambda \sum_{i=1}^T \bm{O}_i\bm{\Delta}_i, \\
    &\forall i.~\bm{O}_i\sim\mathbb{O}(d).
\end{align}
where  $\lambda \in \mathbb{R}^+$ is the scaling coefficient. 
At test time, given task $i$, we retrieve an approximate model $\hat{\bm{\Theta}}_i$ with:
\begin{align}
    \hat{\bm{\Theta}}_i^{RanDeS} &\leftarrow \bm{\Theta}_0 + \lambda \bm{O}_i^{-1}\sum_{i=1}^T \bm{O}_i\bm{\Delta}_i, \nonumber \\
    &= (1-\lambda)\bm{\Theta}_0 + \lambda\bm{\Theta}_i + \lambda\sum_{j\neq i}\bm{O}_i^{-1}\bm{O}_j\bm{\Delta}_j.
\end{align}
Figure~\ref{fig:randes_overview} shows how \emph{RanDeS} works in action with three models.

\subsection{Implication}
% The options are really limitless. We can try flip signs layer-wise, row-wise.
% Any orthogonal transformations can be decomposed into rotation and reflection.
% \hz{benefit of randomness, as well as why it's not practical (huge Os), leads to layer-wise efficient implementations.}
% Another reason for these specfic ones is that arbitrary orthogonal transformations can be decomposed into a permutation and rotation, which correspond to layer shuffling and sign flip respectively. Thus combining them in effect leads to larger range of orthogonal transformations.
\emph{RanDeS} provides a principled framework to reduce interference by decorrelating conflicting deltas into self-cancellation without changing the delta in request. Comparing to model merging~\citep{ilharco2022editing}, \emph{RanDeS} doubles memory footprint to enable task-specific delta retrieval. But it requires effectively \emph{zero} memory overhead to compress more models after the initial setup, since each random orthogonal transformation $\bm{O}_i$ can be specified with a random seed. Thus we can reach higher compression ratio when more models are superposed together.

However, it is impractical to sample the orthogonal transformations directly due to their $O(d^2)$ complexity. We propose two layer-wise implementation to solve this problem: \emph{layer shuffling} and \emph{layer column-wise random sign flips}.

% \begin{figure}[h]
%     \centering
%     \vspace{-4mm}
%     \includegraphics[width=0.5\textwidth]{images/teaser.pdf}
%     \vspace{2mm}
%     \caption{\rbt{Illustration of interference reduction in multi-model compression. \( \bm{M}_0 \) is the pre-trained checkpoint, and \( \bm{M}_i \) the \(i\)-th fine-tuned checkpoint, with task vectors \( \bm{T}_i \), \( \bm{T}_1 \), and \( \bm{T}_2 \). Standard task arithmetic (\( \hat{\bm{M}}_i^{\text{ta}} \), red) sums aligned task vectors, causing interference. With our method (\( \hat{\bm{M}}_i^{\text{our}} \), blue), layer shuffling and superposition decorrelate the interfering task vectors into \( \bm{T}_1' \) and \( \bm{T}_2' \), lowering the Frobenius norm of interference. This allows better retrieval of \( \bm{M}_i \) with a higher merging coefficient \( \lambda \).}}
%     \label{fig:ta-vs-sta}
%     \vspace{-1mm}
% \end{figure}

\subsection{Layer-wise Notation}
Each model $\bm{\Theta}_i$ is a set of parameter matrices:
\[
\bm{\Theta}_i = \left\{\mathbb{I}_i, \left(\bm{M}_i^{k,1}, \bm{M}_i^{k,2}, \dots, \bm{M}_i^{k,m_k} \right)_{k=1}^{K}, \mathbb{O}_i \right\}.
\]
Here $\mathbb{I}_i$ and $\mathbb{O}_i$ are the input and output layers.
Each model has $K$ blocks, with the $k$-th block containing $m_k$ matrices $\bm{M}_i^{k,1}, \bm{M}_i^{k,2}, \dots, \bm{M}_i^{k,m_k}$. We use $\bm{\Delta}_i^{k,1}, \bm{\Delta}_i^{k,2}, \dots, \bm{\Delta}_i^{k,m_k}$ to denote the corresponding delta matrices.

\subsection{Layer Shuffling} \label{sec:layer_shuffle}
\begin{figure}[h]
    \centering
    % \vspace{-4mm}
    \includegraphics[width=0.3\textwidth]{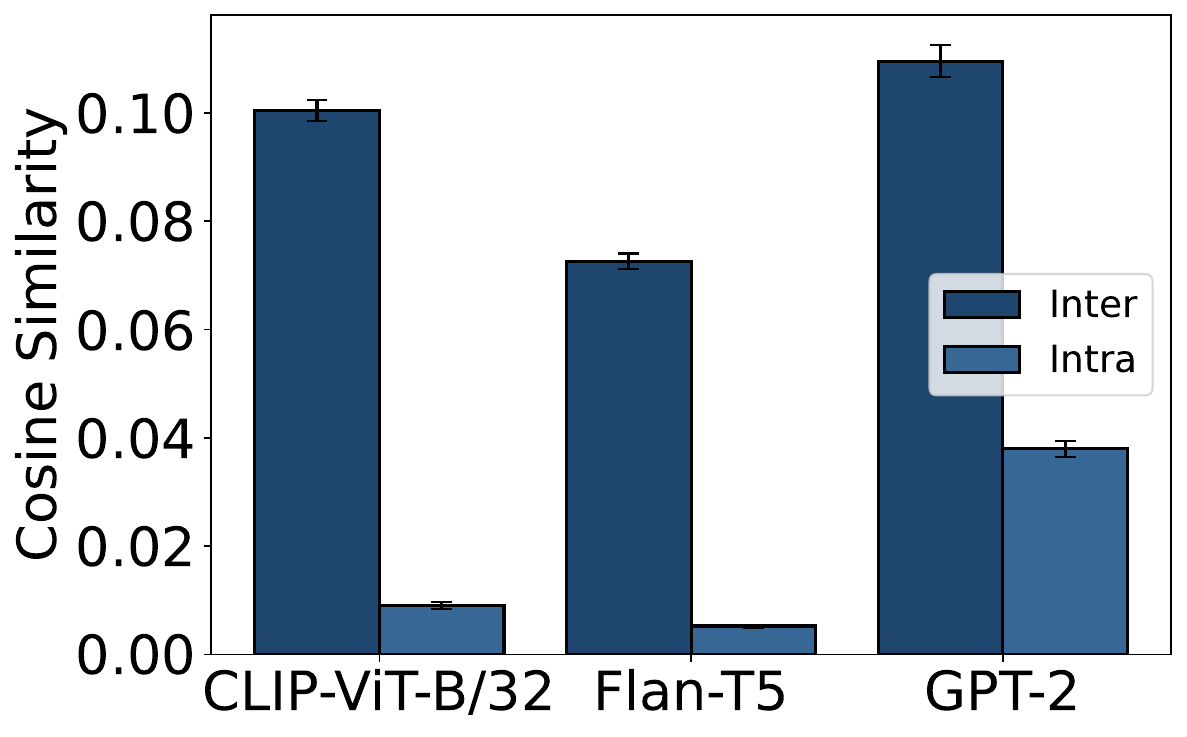}
    % \vspace{2mm}
    \caption{Cosine similarity distributions between delta layers within and across models for CLIP-ViT-B/32, Flan-T5, and GPT-2.}
    \label{fig:layer_cos_sim}
    % \vspace{-1mm}
\end{figure}

% Outline: Shows that mismatching layers is another invertible transformation made possible in offline settings, which can further increase the mutual orthogonality of the merged matrices, leading to potentially smaller interference.
Across several model architectures (CLIP-ViT-B/32~\citep{radford2021learning}, Flan-T5~\citep{longpre2023flan}, and GPT-2~\citep{radford2019language}), we observed that delta layers within the same model exhibit greater variability compared to corresponding layers across fine-tuned models (see Figure~\ref{fig:layer_cos_sim}). 

This suggests that we can decorrelate deltas by shuffling layers within each delta. Leveraging the repeating structure in modern neural networks, for each delta, we \textit{randomly permute} within layers of the same type (e.g., $\{\bm{\Delta}_i^{k,j}\}_{k=1}^K$) to avoid size mismatches. 
Concretely, for each set of layers $\{\bm{\Delta}_i^{k,j}\}_{k=1}^K$ within each delta $\bm{\Delta}_i$, we draw a random permutation $\sigma_i^j\sim \mathbb{S}_K$ from the symmetric group $\mathbb{S}_K$ to shuffle them:
\begin{align}
    &\bm{\Delta}_i^{k,j} \leftarrow \bm{\Delta}_i^{\sigma_i^j(k),j}, \\
    &\bm{\Delta}_i^{Shuffle} = \bm{P}_i\bm{\Delta}_i.
\end{align}
Then, when we want to perform inference using model $\bm{\Theta}_i$, we perform the inverse of the corresponding permutation to obtain the model.
We dub this approach \emph{layer shuffling (RanDeS-S)}. It implements the orthogonal transformation $\bm{O}_i$ as a permutation $\bm{P}_i$, which can be implemented efficiently by reallocating the pointer to each layer.

\subsection{Layer Column-wise Random Sign Flips}\label{sec:sign_flip}
We take inspiration from \citet{cheung2019superposition} on continual learning and introduce \emph{layer column-wise random sign flips (RanDeS-RSF)} as another memory-efficient implementation for the random orthogonal transformations.

Considering merging layer $\bm{\Delta}^{k,j}$ across deltas: $\{\bm{\Delta}_i^{k,j}\in\mathbb{R}^{m,n}\}_{i=1}^T$, we sample random binary diagonal matrices (\emph{context matrices}) whose diagonal entries have equal probability to be $+1$ or $-1$ to each of the $T$ matrices and apply them before summation:
\begin{equation}
    \bm{\Delta}_\star^{k,j} \leftarrow \sum_{i=1}^T\bm{\Delta}_i^{k,j}\bm{D}_i^{k,j}.
\end{equation}
When performing task $i$, we apply the layer-wise inverse transformation $\bm{D}_i^{k,j(-1)}$ to retrieve the delta layer $\bm{\Delta}_i^{k,j}$ from the superposition:
\begin{align}
    \hat{\bm{\Delta}}_i^{k,j}&=\bm{\Delta}_\star^{k,j}\bm{D}_i^{k,j(-1)}, \\
    &= \bm{\Delta}_i^{k,j} + \sum_{l\neq i}[\bm{\Delta}_l^{k,j}\bm{D}_l^{k,j}\bm{D}_i^{k,j(-1)}].
\end{align}
We name this method \emph{layer column-wise random sign flips}, as it decorrelates deltas by randomly flipping the sign for each column. Since $n\ll d$, the layer-wise transformations $\{\bm{D}_i^{k,j}\}_{k,j}$ requires far less memory footprint comparing to the orthogonal transformation $\bm{O}_i$.
\section{Experiments}
We evaluate \emph{RanDeS} across vision and language tasks, showing comparable performance for discriminative and generative models, as well as PEFT models. Through ablation studies, we analyze component importance, merging coefficient effects, and context matrix designs.

\subsection{Experiment Setup}
\begin{figure*}[h]
    \centering
    \includegraphics[width=0.8\textwidth]{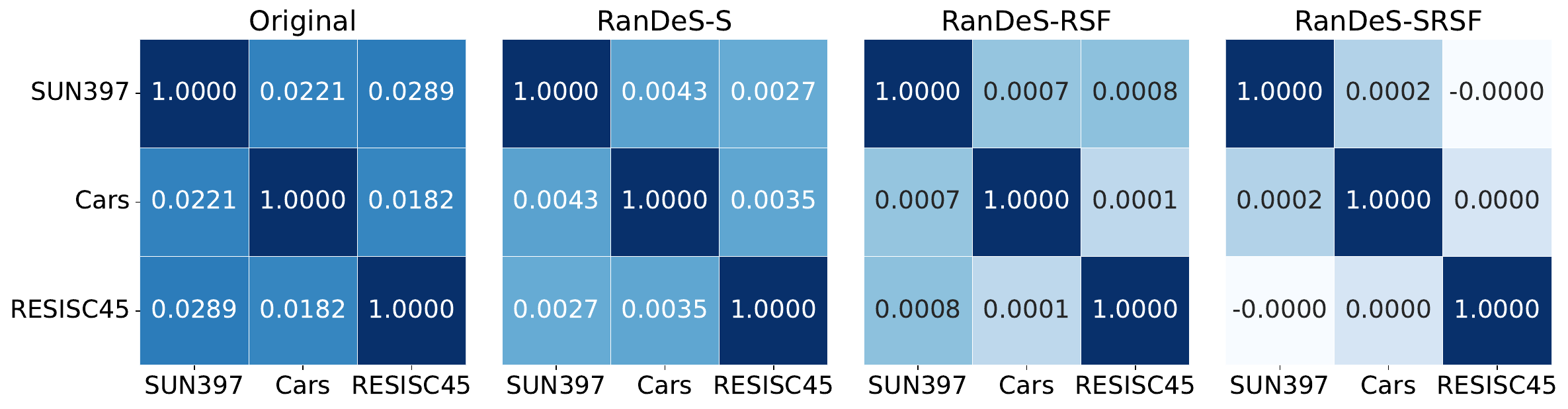}
    \caption{Average pairwise cosine similarity of three out of eight CLIP-ViT-B/32 task vectors during model retrieval for SUN397 across three repetitions. Both \emph{RanDeS-S} and \emph{RanDeS-RSF} increase mutual orthogonality, with an additive effect when combined.}
    \label{fig:pairwise_consine_similarity_sta_clip32}
    \vspace{-1mm}
\end{figure*}
\paragraph{Datasets and Models.}
We follow~\citet{tang2024fusionbench} and select three representative scenarios to evaluate our methods. This includes i) CLIP-ViT-B/32 fine-tuned on eight image classification datasets \rbt{(adopted from~\citep{ilharco2022editing})}; ii) Flan-T5-base fine-tuned on eight text generation datasets; and iii) GPT-2 fine-tuned on seven text classification datasets. Detailed information on the datasets and models is in \hyperref[exp_setup]{Appendix~\ref*{exp_setup}}.

\paragraph{Baselines and Metrics.}
\rbt{
We evaluate baselines from model merging/compression literature, grouped by memory requirements: methods using the original footprint (pre-trained model, standard merging techniques) and those requiring additional memory (fine-tuned models, newer merging baselines). The pre-trained and fine-tuned models provide lower and upper performance bounds respectively. Following~\citet{ilharco2022editing}, we optimize merging coefficient $\lambda$ via validation set grid search. We report accuracy and memory usage across three runs per experiment with random operations. See \hyperref[exp_setup]{Appendix~\ref*{exp_setup}} for details.
}
\subsection{Performance Analysis}
\begin{table*}[h]
    % \caption{Performance comparison on eight image classification tasks with CLIP-ViT-B/32 models merging and compression, \rbt{along with} memory footprint estimate\rbt{s}. 
    % \rbt{In the \textit{Avg.(\%)} column, the values in parentheses indicate performance relative to the fine-tuned model (set to 100\%). In the \textit{Bits(Gb)} column, the values in parentheses represent the relative model size compared to the pre-trained model (set to 1.00).}
    % Three repetitions are done for methods with randomness\rbt{; v}ariances smaller than $0.1\%$ are omitted.}
    
    \caption{\rbt{Performance and memory comparison of CLIP-ViT-B/32 models across eight image classification tasks, showing absolute and normalized accuracy (\%), as well as memory footprint (Gb). Results averaged over three runs where applicable. Variances smaller than 0.1\% are omitted.}}
    \label{clip32-table}
    \begin{center}
        \resizebox{\linewidth}{!}{
            \begin{tabular}{l|cc|*{8}{c}}
                \toprule
                \multicolumn{1}{c}{Method} & \multicolumn{1}{c}{Avg.(\%) \(\uparrow\)} & \multicolumn{1}{c}{Bits(Gb) $\downarrow$} & \multicolumn{1}{c}{SUN397} & \multicolumn{1}{c}{Cars} & \multicolumn{1}{c}{RESISC45} & \multicolumn{1}{c}{EuroSAT} & \multicolumn{1}{c}{SVHN} & \multicolumn{1}{c}{GTSRB} & \multicolumn{1}{c}{MNIST} & \multicolumn{1}{c}{DTD} \\
                \midrule
\rowcolor{gray!15}Pre-trained   & 48.2 \scriptsize{(53.4)} & 0.564 \scriptsize{(1.00)} & 63.2 & 59.8 & 60.7 & 46.0 & 31.6 & 32.5 & 48.3 & 43.9 \\
Weight Averaging   & 66.5 \scriptsize{(73.6)} & 0.564 \scriptsize{(1.00)} & 65.4 & 62.6 & 70.8 & 76.9 & 64.5 & 54.9 & 86.3 & 50.9 \\
Fisher Merging     & 70.6 \scriptsize{(78.2)} & 0.564 \scriptsize{(1.00)} & 66.7 & 64.0 & 72.2 & 91.6 & 69.0 & 64.3 & 83.5 & 53.7 \\
RegMean           & 80.5 \scriptsize{(89.1)} & 0.564 \scriptsize{(1.00)} & 67.8 & 68.9 & 82.5 & 94.4 & 90.6 & 79.2 & 94.7 & 63.2 \\
Task Arithmetic    & 69.8 \scriptsize{(77.2)} & 0.564 \scriptsize{(1.00)} & 64.4 & 61.5 & 70.5 & 80.4 & 73.9 & 62.8 & 93.0 & 51.6 \\
Ties-Merging      & 72.2 \scriptsize{(80.0)} & 0.564 \scriptsize{(1.00)} & 67.1 & 64.2 & 74.1 & 91.6 & 77.7 & 69.4 & 94.1 & 54.0 \\
\rbt{Layerwise} AdaMerging        & 82.6 \scriptsize{(91.5)} & 0.564 \scriptsize{(1.00)} & 67.9 & 71.3 & 83.5 & 92.7 & 87.4 & 92.9 & 98.2 & 67.0 \\
\midrule
\rowcolor{gray!15}Fine-tuned   & 90.3 \scriptsize{(100)} & 2.84 \scriptsize{(5.03)} & 75.0 & 78.3 & 95.2 & 99.0 & 97.3 & 98.9 & 99.6 & 79.7 \\
WEMoE     & 89.2 \scriptsize{(98.8)} & 2.27 \scriptsize{(4.03)} & 73.7 & \underline{76.8} & 93.4 & 98.2 & 96.8 & 98.2 & \textbf{99.6} & 76.6 \\
SMILE  & 89.3 \scriptsize{(98.9)} & 1.23 \scriptsize{(2.\rbt{20})} & 73.6 & \textbf{77.8} & 92.0 & \underline{98.3} & 96.9 & 98.1 & \textbf{99.6} & \underline{78.1} \\
\textbf{RanDeS-S (Ours)} & 81.3 \scriptsize{(90.0)} & \rbt{0.89}  \scriptsize{\rbt{(1.58)}} & 65.6 & 58.5 & 86.8 & 94.5 & 93.2 & 91.4 & 98.5 & 62.2\\
\textbf{RanDeS-RSF (Ours)} & \underline{89.6} \scriptsize{(\underline{99.2})} & \rbt{0.89}  \scriptsize{\rbt{(1.58)}} & \underline{74.4} & 75.6 & \underline{94.6} & \textbf{99.0} & \underline{97.1} & \underline{98.5} & \underline{99.5} & 77.8 \\
\textbf{RanDeS-SRSF (Ours)} & \textbf{89.9} \scriptsize{\textbf{(99.6)}} & \rbt{0.89}  \scriptsize{\rbt{(1.58)}} & \textbf{74.8} & 76.7 & \textbf{94.8} & \textbf{99.0} & \textbf{97.2} & \textbf{98.6} & \underline{99.5} & \textbf{78.7} \\
                \bottomrule
            \end{tabular}
        }
    \end{center}
\end{table*}\label{tab:exp_clip32}
\begin{table*}[h]
    % \caption{Performance comparison on eight GLUE text generation tasks with Flan-T5-base models merging and compression, \rbt{along with} memory footprint estimate\rbt{s}. 
    % \rbt{In the \textit{Avg.(\%)} column, the values in parentheses indicate performance relative to the fine-tuned model (set to 100\%). In the \textit{Bits(Gb)} column, the values in parentheses represent the relative model size compared to the pre-trained model (set to 1.00).}
    % Three repetitions are done for methods with randomness\rbt{; v}ariances smaller than $0.1\%$ are omitted.}
    \caption{\rbt{Performance and memory comparison of Flan-T5-base models across eight GLUE text generation tasks, showing absolute and normalized accuracy (\%), as well as memory footprint (Gb). Results averaged over three runs where applicable. Variances smaller than 0.1\% are omitted.}}
    \label{flan-t5-table}
    \begin{center}
        \resizebox{\linewidth}{!}{
            \begin{tabular}{l|cc|*{8}{c}}
                \toprule
                \multicolumn{1}{c}{Method} & \multicolumn{1}{c}{Avg.(\%) \(\uparrow\)} & \multicolumn{1}{c}{Bits(Gb) $\downarrow$} & \multicolumn{1}{c}{CoLA} & \multicolumn{1}{c}{MNLI} & \multicolumn{1}{c}{MRPC} & \multicolumn{1}{c}{QNLI} & \multicolumn{1}{c}{QQP} & \multicolumn{1}{c}{RTE} & \multicolumn{1}{c}{SST2} & \multicolumn{1}{c}{STSB} \\
                \midrule
\rowcolor{gray!15}Pre-trained & 75.7 \scriptsize{(87.6)} & 1.19 \scriptsize{(1.00)} & 69.1 & 56.5 & 76.2 & 88.4 & 82.1 & 80.1 & 91.2 & 62.2 \\
Weight Averaging  & 78.9 \scriptsize{(91.3)} & 1.19 \scriptsize{(1.00)} & 69.1 & 62.6 & 79.4 & 89.8 & 83.9 & 81.2 & 91.7 & 73.2 \\
Task Arithmetic    & 79.6 \scriptsize{(92.1)} & 1.19 \scriptsize{(1.00)} & 69.7 & 64.1 & 79.2 & 90.2 & 83.9 & 81.6 & 92.1 & 76.4 \\
Ties-Merging      & 79.9 \scriptsize{(92.5)} & 1.19 \scriptsize{(1.00)} & 70.3 & 65.0 & 78.9 & 90.2 & 83.5 & 81.6 & 91.7 & 78.3 \\
\midrule
\rowcolor{gray!15}Fine-tuned         & 86.4 \scriptsize{(100)} & 9.52 \scriptsize{(8.00)} & 75.0 & 83.4 & 87.5 & 91.5 & 85.4 & 85.9 & 93.6 & 88.7 \\
SMILE & 85.5 \scriptsize{(99.0)} & 1.81 \scriptsize{(1.52)} & 73.2 & \textbf{84.2} & 85.0 & 91.3 & \underline{84.9} & \underline{84.8} & \underline{93.5} & 87.3 \\
\textbf{RanDeS-S (Ours)}  & 85.7 \scriptsize{(99.0)} & 2.38 \scriptsize{(2.00)} & 75.5 & 82.0 & 87.5 & 91.1 & 83.9 & 83.8 & \textbf{93.6} & 88.4 \\
\textbf{RanDeS-RSF (Ours)}   & \textbf{86.5} \scriptsize{(100)} & 2.38 \scriptsize{(2.00)} & \textbf{77.2} & 82.1 & \underline{87.6} & \underline{91.6} & \textbf{85.3} & \textbf{85.7} & 93.2 & \textbf{89.0} \\
\textbf{RanDeS-SRSF (Ours)}  & \underline{86.4} \scriptsize{(100)} & 2.38 \scriptsize{(2.00)} & \underline{75.6} & \underline{82.8} & \textbf{88.2} & \textbf{91.7} & \textbf{85.3} & \textbf{85.7} & \underline{93.5} & \underline{88.9} \\
                \bottomrule
            \end{tabular}
        }
    \end{center}
\end{table*}\label{tab:exp_flan-t5}
\paragraph{Superior MTL Performance.} 
\rbt{
Our approach achieves significant accuracy gains across benchmarks (\Cref{clip32-table,flan-t5-table,gpt2-table}), with \textit{RanDeS-SRSF} nearly matching individual fine-tuned models. We outperform WEMoE~\citep{tang2024merging} and SMILE~\citep{tang2024smile} on image classification while using only 40\% and 72\% of their respective memory footprints, and surpass SMILE's text generation performance at benchmark saturation. Though Task Arithmetic~\citep{ilharco2022editing} uses 55\% of our storage, its performance is substantially lower.
}
% We see significant performance improvement in terms of average accuracy across all three benchmarks, per \Cref{clip32-table,flan-t5-table,gpt2-table}. Especially for the image classification tasks (with CLIP-ViT-B/32) and the text generation tasks (with Flan-T5-base), where \textit{RanDeS-SRSF} is approaching the performance of individual fine-tuned models. Among methods that use extra parameters, we surpass both WEMoE~\citep{tang2024merging} and SMILE~\citep{tang2024smile} on image classification despite having lower memory cost. We also outperform SMILE on text generation tasks with higher memory consumption. Meanwhile, Parameter Superposition (PSP)~\citep{cheung2019superposition} performs well in continual learning settings by directly superposing the entire model. But they perform poorly across all the benchmarks . This shows that superposing task vectors is a key ingredient in effective model compression/merging in offline settings.
\paragraph{Amortizable Memory Overhead.} 
\rbt{
Our method requires only 2x memory, mainly from storing the delta superposition. By storing random seeds and regenerating transformations on-the-fly with minimal overhead (292.70 ms for CLIP-ViT-B/32, 658.19 ms for CLIP-ViT-L/14 on Intel Xeon Gold 6448Y CPU), we achieve effectively zero additional memory per model. This enables efficient scaling, demonstrated by merging 20 CLIP-ViT-L/14 models with state-of-the-art performance and 9x memory reduction (Sec.~\ref{sec:scale}).
}
% Note that our method only doubles the memory footprint, despite the large number of tasks learned. The extra storage cost comes from the merged task vectors $\{\bm{T}^k_\star\}_{k=1}^K$, as well as the context matrices $\{\bm{C}^k_\star\}_{k=1}^K$, as mentioned in equation~\ref{param_shuffle} and \ref{param_superpose}. The merged task vectors contributes the majority of additional storage cost ($>99\%$), as the context matrices \rbt{consist of binary numbers}, which can be stored very efficiently. Moreover, we can choose to store only the random seeds, and regenerate the corresponding context matrices and layer shuffling orders on-the-fly. \rbt{This introduces minimal forward pass overhead (292.70 ms for retrieving a CLIP-ViT-B/32 model and 658.19 ms for CLIP-ViT-L/14 models on an Intel Xeon Gold 6448Y CPU) and} leads to effectively \textit{zero} memory overhead for every additional model in compression. \rbt{This process will amortize the memory footprint when more models are merged together, as demonstrated in Sec.~\ref{sec:scale} where we merge 20 CLIP-ViT-L/14 models with state-of-the-art performance and a ninefold memory reduction.}

\subsection{Key Components Ablation}
\label{sec:key-comp-ablation}
In this section, we ablate both the \textit{layer shuffling} and \textit{layer column-wise random sign flips} to show their individual contribution.

As shown in \Cref{clip32-table,flan-t5-table,gpt2-table,flan-t5-lora-table,scale}, both methods significantly outperform task arithmetic consistently.
In some benchmarks, shuffling works better (Flan-T5-base and GPT-2) while random sign flips works better in others (CLIP).
This difference may be because of the nature of deltas themselves: how they vary across the model layers and across different models.
We find that the combined approach is able to combine gains from both components, yielding consistently the best result across all the benchmarks. This complementary effect is further manifested in Figure~\ref{fig:pairwise_consine_similarity_sta_clip32}, where introducing both mechanism gets the smallest pairwise cosine similarity among interfering deltas.
\rbt{We provide a more detailed analysis of the interplay between shuffling and random sign flips in Section~\ref{sec:extra_analysis}.}

%, with inconsistent ranking depending on the nature of models being compressed. Interestingly, \textit{TA+Mismatch} and \textit{STA} surpass the combined approach on Flan-T5-base and GPT-2 by $0.1\%$. This is likely due to the variance coming from the method itself, as well as the coarse-grained grid search of merging coefficient $\lambda$.

\subsection{Impact of Merging Coefficient $\lambda$}

\begin{figure}[h]
    \centering
    \includegraphics[width=0.4\textwidth]{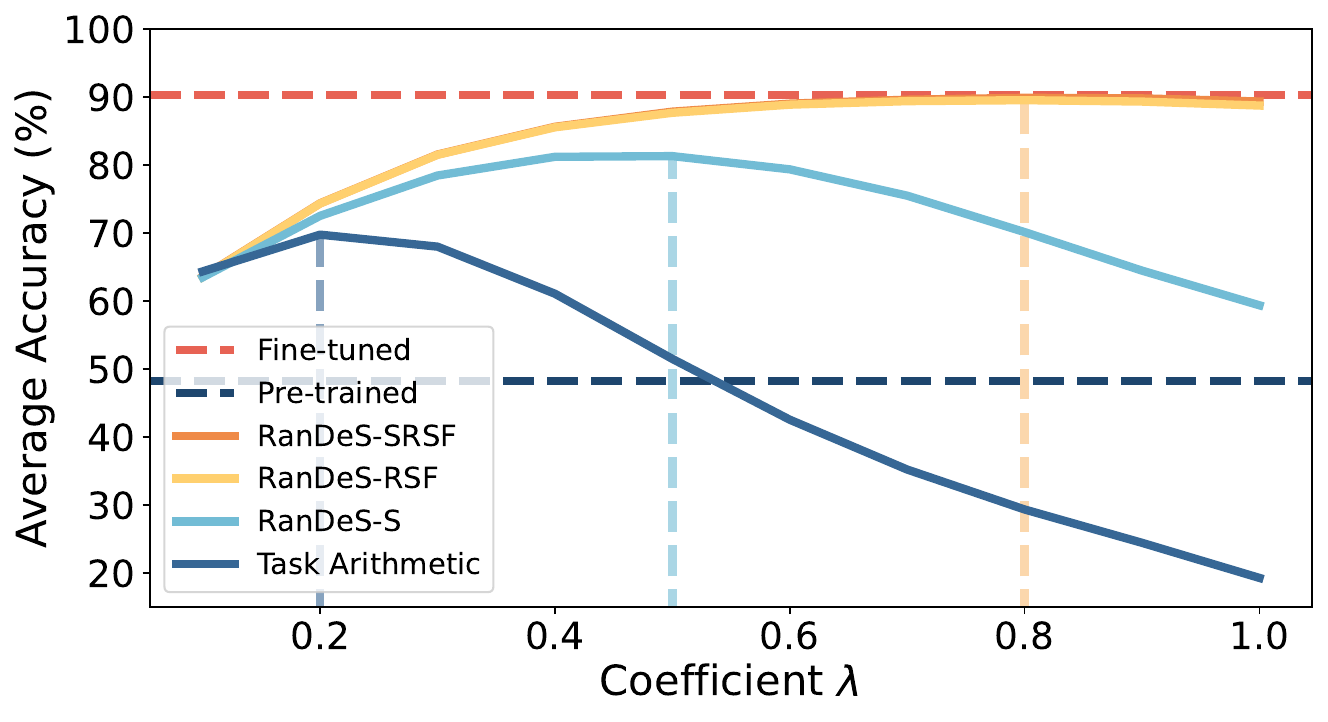}
    \caption{The impact of $\lambda$ on average accuracy over eight image classification tasks.}
    % \vspace{-10mm}
    \label{fig:clip32_avg_acc_lambda}
\end{figure}

% \begin{figure}{r}{0.2\textwidth}
%     \centering
%     % \vspace{-16mm}
%     \includegraphics[width=0.18\textwidth]{images/clip32_avg_acc_lambda.pdf}
%     \caption{The impact of $\lambda$ on average accuracy over eight image classification tasks.}
%     % \vspace{-10mm}
%     \label{fig:clip32_avg_acc_lambda}
% \end{figure}

Here we examine the interplay between the merging coefficient $\lambda$ and the average performance across different setup. For each variant, we perform a grid search on $\lambda=\{0.1, 0.2, \cdots, 1.0\}$ when compressing eight CLIP-ViT-B/32 models for image classification.
\hyperref[fig:clip32_avg_acc_lambda]{Figure~\ref*{fig:clip32_avg_acc_lambda}} shows the change of optimal model performance and the coefficient $\lambda$ when \textit{layer shuffling} and \textit{column-wise random sign flips} are introduced to task arithmetic.

We observe that when shuffling and random sign flips are introduced, the best performance increases along with the value of $\lambda$. This shows the effectiveness of our method in reducing interference, allowing larger $\lambda$ to be selected for more authentic model retrieval.

\subsection{Impact of Context Matrix Design} 
% \textbf{Goal}: i) Show accuracy vs. memory footprint; ii) Reveal what makes a context works.

To further examine how \textit{RanDeS-RSF} works and shad light on better context matrix design, we make a comparison between three types of context matrices: random binary diagonal matrix with $\{-1, +1\}$ entries (RBD), identity matrix (Identity), and random diagonal matrix with entries draw from Normal distribution (RD). We use random layer shuffling when compressing the 8 CLIP-ViT-B/32 models on the image classification tasks. 

\begin{figure}[h]%
\centering     %%% not \center
\subfigure[Context Matrix Design]{\includegraphics[width=5cm]{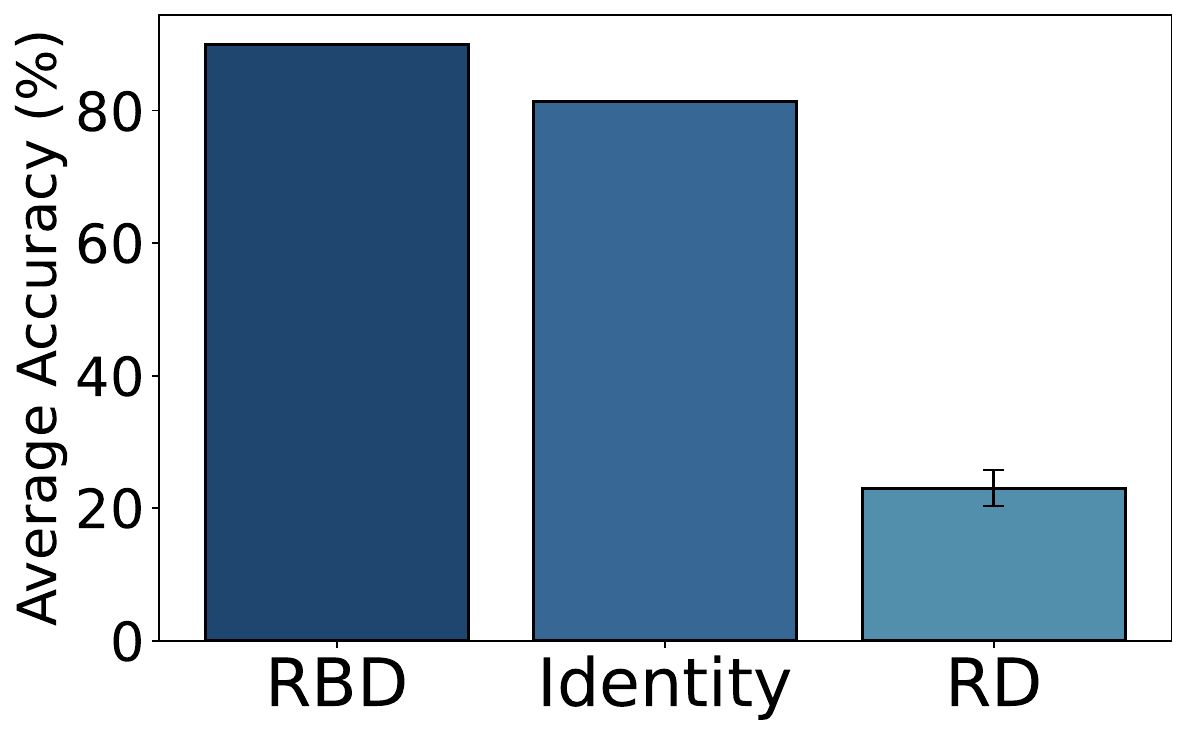}}%
\hspace{1cm}
\subfigure[Target Layer Selection]{\includegraphics[width=5cm]{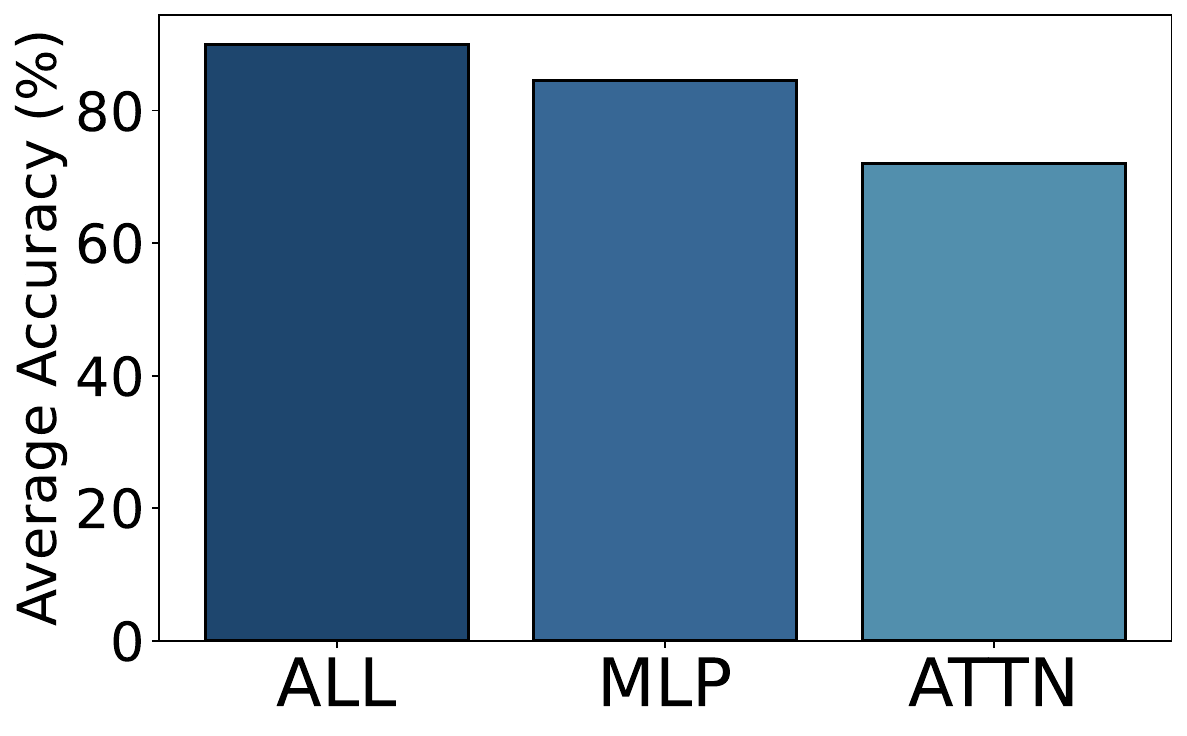}}%
\caption{(a) Impact of context matrix design to the average accuracy. RBD stands for random binary diagonal; RD stands for random diagonal. (b) Impact of target layer selection to the average accuracy. ALL stands for choosing all layers; MLP means only MLP layers are selected; and ATTN stands for attention layers.}%
\label{fig:context_and_target}
% \vspace{-10mm}
\end{figure}

The average accuracy and its variance with the optimal merging coefficient is shown in Figure~\ref{fig:context_and_target}(a). RBD receives higher accuracy than Identity due to the randomness it introduces, which reduces interference. Despite being random, RD's accuracy is much lower than RBD. We think this happens because RD is not an orthogonal matrix. It fails to preserve the magnitude and thus disturb this self-cancellation process.

% three barplot, each with two bars for random binary diagonal matrix and random diagonal matrix. (each has two split, left hand side the average accuracy with same random seed for all layers of each model; right hand side the average accuracy with different random seed for different layers.

% Context types: identity matrix (reduce to task arithmetic), random diagonal binary matrix, random diagonal matrix, random rotation matrix drawn from orthogonal group O(n), and random dense matrix (pseudo-inverse is used for retrieval when the matrix is not invertible).

% One setting where we switch fixed to different matrix across layers. Do it on the winners.

\subsection{Target Layer Selection}
By default we apply the random operations on all layers within the models. In this section, we evaluate the benefits of targeting specific types of layers. To do this, we create two variants: MLP (which selects only the MLP layers) and ATTN (which selects only the attention layers), in addition to the default setup (ALL). Figure~\ref{fig:context_and_target} (b) shows the average accuracy for each setup across eight image classification tasks using CLIP-ViT-B/32. The ALL configuration achieves the highest accuracy, followed by MLP and ATTN. Note that the total number of parameters in MLP is twice that of ATTN, explaining the gradual decline in performance as fewer parameters are selected.

\subsection{Model Hot Swapping}
The ability to hot-swap models in real-world applications is crucial, especially in dynamic environments like model serving, where new models need to be integrated into the system regularly, and deprecated ones need to be removed in a timely fashion. As mentioned, the \textit{RanDeS-SRSF} method allows for this by shuffling layers and sampling diagonal binary matrices \textit{independently} of data or model parameters, thus enabling the on-the-fly addition of new models without the need for recomputation. This provides a big advantage over methods like WEMoE which require recomputation of the router when new models are added~\citep{tang2024merging}\rbt{, or TALL-masks, which needs to recompute all task masks when a new model is added~\citep{wang2024localizing}}.
\begin{table}[h]
\caption{Comparison of selected methods with hot adding and recomputation requirements when new models are added to the pool.}
\label{tab:hot-adding}
\begin{center}
\begin{tabular}{ccc}
\multicolumn{1}{c}{\bf Method} &\multicolumn{1}{c}{\bf Hot Swap} &\multicolumn{1}{c}{\bf Recomputation}
\\ \hline 
Task Arithmetic     & \checkmark   & \ding{55}  \\
WEMoE         & \ding{55}       & \checkmark \\
TALL-masks  & \ding{55}       & \checkmark \\
RanDeS-SRSF           & \checkmark   & \ding{55} \\
\end{tabular}
\end{center}
\end{table}
\subsection{Parameter Efficient Finetuning (PEFT) Model Compression}
We can also apply our method on PEFT adapter weights. Consider LoRA \citep{hu2021lora}, where we have a fixed pre-trained model $\bm{\Theta}_0$, along with LoRA weights $\bm{L}_i$. We merge the LoRA weights to get the fine-tuned model: $\bm{\Theta}_i=\bm{\Theta}_0+\rbt{\lambda}\bm{L}_i$. We apply \textit{layer shuffling} and \textit{random column-wise sign flips} on these LoRA weight vectors before retrieval. 

Experiments on Flan-T5-base LoRA fine-tunes \citep{longpre2023flan,tang2024fusionbench,tang2024smile} demonstrate that our method is performative in PEFT compression settings \rbt{as well} (Table~\ref{flan-t5-lora-table}). With \rbt{99.8\%} normalized average accuracy compare to the fine-tuned baseline, and 1.20 Gb memory usage, our method presents a \rbt{better} trade-off point between performance and storage usage \rbt{than the state-of-the-art model SMILE~\citep{tang2024smile}}.

\begin{table*}[h]
    % \caption{Multi-task performance when merging Flan-T5-base LoRA models on eight GLUE tasks, \rbt{along with} memory footprint estimate\rbt{s}. 
    % \rbt{In the \textit{Avg.(\%)} column, the values in parentheses indicate performance relative to the fine-tuned model (set to 100\%). In the \textit{Bits(Gb)} column, the values in parentheses represent the relative model size compared to the pre-trained model (set to 1.00).}
    % Three repetitions are done for methods with randomness\rbt{; v}ariances smaller than $0.1\%$ are omitted.}
    \caption{\rbt{Performance and memory comparison of Flan-T5-base LoRA models across eight GLUE text generation tasks, showing absolute and normalized accuracy (\%), as well as memory footprint (Gb). Results averaged over three runs where applicable. Variances smaller than 0.1\% are omitted.}}
    \label{flan-t5-lora-table}
    \begin{center}
        \resizebox{\linewidth}{!}{
            \begin{tabular}{l|cc|*{8}{c}}
                \toprule
                \multicolumn{1}{c}{Method} & \multicolumn{1}{c}{Avg.(\%) \(\uparrow\)} & \multicolumn{1}{c}{Bits(Gb) $\downarrow$} & \multicolumn{1}{c}{CoLA} & \multicolumn{1}{c}{MNLI} & \multicolumn{1}{c}{MRPC} & \multicolumn{1}{c}{QNLI} & \multicolumn{1}{c}{QQP} & \multicolumn{1}{c}{RTE} & \multicolumn{1}{c}{SST2} & \multicolumn{1}{c}{STSB} \\
                \midrule
\rowcolor{gray!15}Pre-trained & 75.7 \scriptsize{(87.6)} & 1.19 \scriptsize{(1.00)} & 69.1 & 56.5 & 76.2 & 88.4 & 82.1 & 80.1 & 91.2 & 62.2 \\
% \midrule
Weight Averaging  & 78.2 \scriptsize{(92.4)} & 1.19 \scriptsize{(1.00)} & \textbf{69.7} & 59.7 & 78.9 & 90.1 & 83.8 & 80.5 & 91.2 & 72.0 \\
Task Arithmetic    & 77.4 \scriptsize{(91.5)} & 1.19 \scriptsize{(1.00)} & 68.8 & 55.2 & 78.7 & 89.8 & 83.7 & 79.1 & 91.5 & 72.4 \\
Ties-Merging   & 77.5 \scriptsize{(91.6)} & 1.19 \scriptsize{(1.00)} & 68.3 & 56.3 & 79.4 & 89.8 & 83.7 & 79.4 & 91.6 & 71.2 \\
\midrule
\rowcolor{gray!15}Fine-tuned         & 84.6 \scriptsize{(100)} & 1.25 \scriptsize{(1.05)} & 69.1 & 82.7 & 85.5 & 90.9 & 84.0 & 84.4 & 92.9 & 87.4 \\
SMILE & \underline{84.0} \scriptsize{(\underline{99.3})} & 1.21 \scriptsize{(1.02)} & \underline{69.3} & \textbf{82.9} & 83.8 & \underline{90.6} & \underline{83.9} & 83.4 & \textbf{93.1} & 85.1 \\
% \midrule
\rbt{\textbf{RanDeS-S (Ours)}}   & \rbt{83.9 \scriptsize{(99.2)}} & 1.20 \scriptsize{(1.01)} & \rbt{69.2} & $\rbt{79.0} \pm \rbt{0.3}$ & \rbt{\underline{84.2}} & \rbt{90.4} & \rbt{\textbf{84.1}} & \rbt{\textbf{85.0}} & \rbt{\underline{92.9}} & \rbt{\underline{86.5}} \\
\rbt{\textbf{RanDeS-RSF (Ours)}}   & \rbt{83.0} \scriptsize{\rbt{(98.1)}} & 1.20 \scriptsize{(1.01)} & \rbt{69.1} & \rbt{81.3} & \rbt{82.2} & \rbt{90.5} & \rbt{83.2} & \rbt{79.1} & \rbt{92.7} & \rbt{85.6} \\
\textbf{\rbt{RanDeS-SRSF (Ours)}}  & \rbt{\textbf{84.4}} \scriptsize{\rbt{(\textbf{99.8})}} & 1.20 \scriptsize{(1.01)} & \rbt{69.1} & \rbt{\underline{82.7}} & \rbt{\textbf{85.0}} & \rbt{\textbf{90.9}} & \rbt{83.8} & \rbt{\underline{84.2}} & \rbt{92.7} & \rbt{\textbf{86.9}} \\
                \bottomrule
            \end{tabular}
        }
    \end{center}
\end{table*}

\rbt{
\subsection{Scalability Analysis}\label{sec:scale}
Our method scales effectively to merging larger models and more tasks, as demonstrated on CLIP ViT-L/14 with 8, 14, and 20 image classification tasks (Table~\ref{scale}). RanDeS-SRSF achieves near fine-tuned performance (93.5\% vs 94.2\% for 20 tasks) while maintaining constant 2.87GB storage regardless of task count. In contrast, the state-of-the-art model TALL-masks+TA~\citep{wang2024localizing} requires progressively more storage (5.42GB to 9.25GB) as tasks increase. Though Task Arithmetic uses only 1.59GB, its performance drops significantly with more tasks. Model retrieval remains efficient, requiring just 658.19ms per CLIP ViT-L/14 model on an Intel Xeon Gold 6448Y CPU.
\begin{table*}[ht]
\centering
\caption{\rbt{Performance and memory comparison of CLIP ViT-L/14 models across three test scenarios with 8, 14, and 20 image classification tasks, showing absolute and normalized accuracy (\%), as well as memory footprint (Gb). Results averaged
over three runs where applicable. Variances smaller than 0.1\% are omitted.}}
\label{scale}
\resizebox{\textwidth}{!}{%
\begin{tabular}{lcccccc}
\toprule
Method & \multicolumn{2}{c}{8 tasks} & \multicolumn{2}{c}{14 tasks} & \multicolumn{2}{c}{20 tasks} \\
\cmidrule(lr){2-3} \cmidrule(lr){4-5} \cmidrule(lr){6-7}
 & Acc.(\%) $\uparrow$ & Bits(Gb) $\downarrow$ & Acc.(\%) $\uparrow$ & Bits(Gb) $\downarrow$ & Acc.(\%) $\uparrow$ & Bits(Gb) $\downarrow$ \\
\midrule
\rowcolor{gray!15}
Pre-trained & 64.5 \scriptsize{(68.3)} & 1.59 \scriptsize{(1.00)} & 68.1 \scriptsize{(72.8)} & 1.59 \scriptsize{(1.00)} & 65.2 \scriptsize{(69.2)} & 1.59 \scriptsize{(1.00)} \\
\addlinespace[0.1cm]
Task Arithmetic & 84.0 \scriptsize{(88.7)} & \textbf{1.59} \scriptsize{(\textbf{1.00})} & 79.1 \scriptsize{(84.2)} & \textbf{1.59} \scriptsize{(\textbf{1.00})} & 73.8 \scriptsize{(78.3)} & \textbf{1.59} \scriptsize{(\textbf{1.00})} \\
\midrule
\rowcolor{gray!15}
Fine-tuned & 94.4 \scriptsize{(100)} & 10.53 \scriptsize{(6.62)} & 93.5 \scriptsize{(100)} & 18.18 \scriptsize{(11.43)} & 94.2 \scriptsize{(100)} & 25.84 \scriptsize{(16.25)} \\
\addlinespace[0.1cm]
Magnitude Masking & 92.8 \scriptsize{(98.2)} & 5.42 \scriptsize{(3.41)} & 90.6 \scriptsize{(96.7)} & 7.34 \scriptsize{(4.62)} & 90.9 \scriptsize{(96.4)} & 9.25 \scriptsize{(5.82)} \\
TALL Mask+TA & 94.2 \scriptsize{(99.7)} & 5.42 \scriptsize{(3.41)} & 92.4 \scriptsize{(98.8)} & 7.34 \scriptsize{(4.62)} & 93.2 \scriptsize{(98.9)} & 9.25 \scriptsize{(5.82)} \\
\textbf{RanDeS-S (Ours)} & 93.0 \scriptsize{(98.4)} & \underline{2.87} \scriptsize{(\underline{1.81})} & 88.8 \scriptsize{(94.6)} & \underline{2.87} \scriptsize{(\underline{1.81})} & $87.1 \pm 0.1$ \text{\scriptsize{(92.2)}} & \underline{2.87} \scriptsize{(\underline{1.81})} \\
\textbf{RanDeS-RSF (Ours)} & \underline{94.2} \scriptsize{(\underline{99.8})} & \underline{2.87} \scriptsize{(\underline{1.81})} & \underline{92.8} \scriptsize{(\underline{99.2})} & \underline{2.87} \scriptsize{(\underline{1.81})} & \underline{93.4} \scriptsize{(\underline{99.1})} & \underline{2.87} \scriptsize{(\underline{1.81})} \\
\textbf{RanDeS-SRSF (Ours)} & \textbf{94.3} \scriptsize{(\textbf{99.9})} & \underline{2.87} \scriptsize{(\underline{1.81})} & \textbf{93.0} \scriptsize{(\textbf{99.5})} & \underline{2.87} \scriptsize{(\underline{1.81})} & \textbf{93.5} \scriptsize{(\textbf{99.3})} & \underline{2.87} \scriptsize{(\underline{1.81})} \\
\bottomrule
\end{tabular}
}
\end{table*}

}

\section{Related Work}
\paragraph{Model Merging.}
Recent research on model merging is largely founded on \textit{linear mode connectivity (LMC)}~\citep{frankle2020linear,neyshabur2020being}, which posits that models fine-tuned from the same pre-trained model are connected by a linear path along which performance remains constant. 
Building upon this concept, \citet{wortsman2022model} and \citet{li2024scalable} demonstrated that a set of specialist models can be directly interpolated to obtain a multi-task model. \citet{ilharco2022editing} proposed interpolating the parameter deltas (referred to as "task vectors") instead. 
However, these methods suffer from \textit{task interference}: when different models adjust the same parameters in conflicting ways, summing these adjustments leads to interference and degraded performance on individual tasks~\citep{yadav2024ties,tang2024smile,wang2024localizing}.
To mitigate this interference, various strategies have been proposed.
\citet{yang2023adamerging} optimized the merging coefficients for different tasks and layers to reduce interference. \citet{yadav2024ties} addressed the conflict by removing redundant parameters and resolving sign disagreements. \citet{tang2024merging} reduced interference by upscaling the multilayer perceptron (MLP) layers. \citet{tang2024smile} compressed task vectors using singular value decomposition (SVD) and performed routing between them to further diminish interference. 
Both \citet{wang2024localizing} and \citet{yu2024language} sparsified the task vectors to prevent task conflicts.
Additionally, \citet{ortiz2024task} proposed fine-tuning the linearized model along the tangent space of the pre-trained model to promote weight disentanglement and avoid interference.
In contrast to the above mentioned methods that aim to avoid conflicts, we intentionally accumulate interference among conflicting task vectors to facilitate their mutual cancellation.

\paragraph{Model Compression.}
Model compression techniques aim to reduce the memory footprint of models while maintaining their performance. Model pruning compresses neural networks by removing inessential parameters in either a structured~\citep{anwar2017structured,fang2023depgraph,he2023structured,wang2019structured} or unstructured~\citep{liao2023can,kwon2020structured} manner.
Parameter quantization saves memory and speeds up inference by converting the weights and activation values of a neural network from high precision to low precision~\citep{gholami2022survey,liu2021post,yuan2022ptq4vit}.
Knowledge distillation reduces the memory footprint by training a smaller network to mimic a larger network's behavior~\citep{gou2021knowledge,cho2019efficacy,park2019relational,zhao2022decoupled}. Leveraging the low-rank nature of model parameters, many works decompose weight matrices into low-rank matrices for memory reduction
~\citep{yu2017compressing,li2023losparse,guo2024learning}. 
\citet{ryu2023efficient} observed the low-rank nature of weight residuals in overparameterized models and proposed reducing storage demands for fine-tuned models through low-rank approximation of these residuals. Similarly, \citet{tang2024smile} compresses individual task vectors using SVD and routes through a set of them conditioned on input. Our work differs from these works in that we try to reduce redundancy across a set of aligned models rather than within them.
\section{Discussion and Future Work}
\rbt{
In this work, we introduce \emph{RanDeS} as a principled framework to reduce interference in model merging frameworks, and two layer-wise efficient implementations.
These data- and model-agnostic random operations enable users to i) efficiently modify the model merging combinations without the need for additional training or optimization; ii) merge additional models without increasing memory usage by saving random seeds.
Evaluation on diverse model and task sets demonstrates that our method maintains high performance while keeping a constant memory footprint as more and larger models are merged.
These attributes make our approach highly practical for real-world multi-model serving environments.
}

\rbt{
An interesting future direction is to further improve performance by increasing orthogonality, potentially through alternative random operations or more systematic approaches. Our method relies on specific properties of model parameters that emerge from fine-tuning. 
Identifying these properties and enhancing fine-tuning strategies could lead to better merging and compression performance.
Since we reduce cross-model redundancy, applying model compression algorithms could potentially further decrease memory footprint.}
% Future research will further refine these strategies, particularly exploring the role of specific layer subsets in model merging and optimizing the randomization processes for improved results.

%\VK{or another way to say it: previous work adjusts the magnitude of the task vectors, but we are going to also adjust the direction of the vectors}

% Can we superpose inputs too? And make superposed models to perform superposed computation.

% Explore the PEFT perspective -- context as LoRA modules.

% Experiment results show that not every task use less significant dimensions during fine-tuning. So SMILE's conclusion is not accurate.

% Tasks that use less significant spaces:
% MRPC, \

% future work: find important layer subset to superpose with, use differentiable context to improve performance.

% \newpage
\bibliographystyle{plainnat}
\bibliography{example_paper}

\newpage
\appendix
\onecolumn

\section{Overview}
In this appendix we present more information about the experiment settings and analysis that could not fit in the main paper. In Sec.~\ref{exp_setup} we present more details about the datasets, models, and baseline methods used in evaluation.
% In Sec.~\ref{sec:memory_estimate} we explain how we estimate the memory footprint for each baseline method. 
In Sec.~\ref{sec:extra_analysis} we include additional analysis and results.
\section{Experiment Setup} \label{exp_setup}
This section provides detailed descriptions for the datasets, baselines, and model fine-tuning settings.
\paragraph{Datasets Details.} 
Evaluation are performed on two sets of datasets with different type of tasks.
\begin{enumerate}
\item \textbf{Image Classification Datasets}: For image classification, following~\citet{ilharco2022editing,tang2024fusionbench,wang2024localizing}, we use twenty tasks from CLIP's~\citep{radford2021learning} test set: SUN397~\citep{sun397}, Cars~\citep{krause20133d}, RESISC45~\citep{cheng2017remote}, EuroSAT~\citep{helber2019eurosat}, SVHN~\citep{netzer2011reading}, GTSRB~\citep{stallkamp2012man}, MNIST~\citep{deng2012mnist}, DTD~\citep{cimpoi14describing}, CIFAR100~\citep{Krizhevsky2009LearningML}, STL10~\citep{coates2011analysis}, Flowers102~\citep{nilsback2008automated}, OxfordIIITPet~\citep{parkhi2012cats}, PCAM~\citep{veeling2018rotation}, FER2013~\citep{goodfellow2013challenges}, EMNIST~\citep{cohen2017emnist}, CIFAR10~\citep{Krizhevsky2009LearningML}, Food101~\citep{bossard2014food}, FashionMNIST~\citep{xiao2017fashion}, RenderedSST2~\citep{socher2013recursive,radford2019language}, and KMNIST~\citep{clanuwat2018deep}. For experiments on K tasks, the first K datasets from this list are select.
\item \textbf{Text Classification and Generation Datasets}: For text classification and generation, following~\citet{tang2024fusionbench}, we have in total eight tasks from the GLUE benchmark~\citep{wang2018glue}: CoLA, MNLI, MRPC, QNLI, QQP, RTE, SST2, and STSB.
\end{enumerate}
\paragraph{Baseline Details.} Our experiments compare the following baselines and our methods:
\begin{itemize}
    \item \textbf{Pre-trained}: Pre-trained model used across all tasks (performance lower bound).
    \item \textbf{Fine-tuned}: Individual fine-tuned models (performance upper bound).
    \item \textbf{Weight Averaging}~\citep{wortsman2022model}: Merge models by directly averaging their parameters.
    \item \textbf{Fisher Merging}~\citep{matena2022merging}: Fisher Merging uses the Fisher information as a weight for each parameter during weight averaging.
    \item \textbf{RegMean}~\citep{jin2022dataless}: RegMean introduces a constraint in model merging by minimizing the L2 distance between the merged model and each individual model.
    \item \textbf{Task Arithmetic}~\citep{ilharco2022editing}: Task Arithmetic computes the delta parameters between fine-tuned models and the base model (known as "task vectors") and aggregates them before adding into a pre-trained model.
    \item \textbf{Ties-Merging}~\citep{yadav2024ties}: Ties-Merging addresses task conflict issues found in Task Arithmetic by eliminating redundant parameters and resolving symbol conflicts.
    \item \textbf{Layer-wise AdaMerging}~\citep{yang2023adamerging}: Layer-wise AdaMerging finds optimal merging coefficients for each layer of each task vector in Task Arithmetic using test-time adaptation.
    \item \textbf{WEMoE}~\citep{tang2024merging}: WEMoE only merges the layer norm and attention layers while keeping the multi-layer perceptron layers unmerged, with a router to dynamically allocate weights to each MLP conditioned on the input. 
    \item \textbf{SMILE}~\citep{tang2024smile}: SMILE compresses task vectors with singular value decomposition (SVD). It then determines the routing weights based on the alignment between input and each low-rank matrix.
    \item \textbf{TALL-masks+TA}~\citep{wang2024localizing}: TALL-masks+TA finds a binary parameter mask for each task vector by finding task-specific parameters with values deviate a lot from the aggregated multi-task vector. The corresponding mask for each task is applied to the multi-task vector before adding to a pre-trained model.
    \item \textbf{Magnitude Masking}~\citep{wang2024localizing}: Magnitude Masking differ from TALL-masks in that it determines per-task masks by keeping the top k\% of each task vector's parameters.
    \item \textbf{RanDeS-S (Ours)}: \emph{RanDeS-S} performs layer shuffling among the repetitive layers in each delta before merging them with Task Arithmetic.
    \item \textbf{RanDeS-RSF (Ours)}: \emph{RanDeS-RSF} applies random column-wise sign flips to each layer in each delta in Task Arithmetic.
    \item \textbf{RanDeS-SRSF (Ours)}: \emph{RanDeS-SRSF} combines layer shuffling and column-wise random sign flips.
\end{itemize}
\paragraph{Model Details.} 
We utilize fine-tuned models from~\citet{tang2024fusionbench} and \citet{wang2024localizing}. Here we describe the experimental setup for fine-tuning these models.
\begin{itemize}
    \item \textbf{CLIP-ViT-B/32 Models}: The CLIP-ViT-B/32 models are fine-tuned by~\citet{tang2024fusionbench}. The Adam optimizer is employed with a fixed learning rate of 1$e^{-5}$ for a total of 4,000 training steps with the batch size of 32. The zero-shot classification layer is computed on-the-fly with a frozen text encoder.
    \item \textbf{CLIP-ViT-L/14 Models}: Different from CLIP-ViT-B/32 models, these models are fine-tuned by~\citet{wang2024localizing} with the training procedure described in~\citet{ilharco2022editing}. The AdamW optimizer is employed with a fixed learning rate of 1$e^{-5}$ for a total of 2,000 training steps with the batch size of 128, and a cosine annealing learning rate schedule with 200 warm-up steps. The zero-shot classification heads are pre-computed and freezed during fine-tuning process, following~\citet{ilharco2022editing} and \citet{ortiz2024task}.
    \item \textbf{GPT-2 Models}: These models are fine-tuned by~\citet{tang2024fusionbench} with a constant learning rate of 5$e^{-5}$ for 3 epochs.
    \item \textbf{Flan-T5-base and LoRA Models}: These models come from~\citet{tang2024fusionbench}, with unspecified fine-tuning settings.
\end{itemize}
\paragraph{Evaluation Metrics.}
We measure performance using average task accuracy and normalized accuracy (relative to Fine-tuned baseline). For STSB~\citep{wang2018glue}, we use Spearman's correlation. Memory efficiency is evaluated by estimated memory footprint in Gb and normalized footprint (relative to Pre-trained baseline).
% , detailed in Sec.~\ref{sec:memory_estimate}.
\paragraph{Default Experimental Setup.}
We use global random seeds 42, 43, 44 for three runs per experiment on random approaches. Each model's specific seed is generated by adding its index to the global seed, and is used consistently for layer shuffling and binary diagonal matrices across target layers. Following~\citet{ilharco2022editing}, we apply uniform merging coefficients across models, optimized via grid search on validation sets (10\% of training data, max 1,000 samples~\citep{wang2024localizing}). The search space is ${0.1, 0.2, \cdots, 1.0}$, extended to ${0.1, 0.2, \cdots, 2.0}$ for Flan-T5-base LoRA experiments.

% \section{Analysis on Other Model Merging Frameworks}\label{sec:other_merging_frameworks}

% \section{Derivation of \hyperref[interference_frob_norm_equation]{Equation~\ref*{interference_frob_norm_equation}}} \label{interference_frob_norm_derivation}
% Here we derive the squared Frobenius norm of the interference $\lambda\sum_{i\neq k}\bm{T}_i^k$ in more details:
% \begin{align}
%     \left\|\lambda\sum_{i\neq k}\bm{T}_i^k\right\|_F^2 &= \left\langle\lambda\sum_{i\neq k}\bm{T}_i^k,\lambda\sum_{i\neq k}\bm{T}_i^k\right\rangle_F, \\
%     &=\lambda^2\left(\sum_{i\neq k}\sum_{j\neq k}\langle\bm{T}_i^k, \bm{T}_j^k\rangle_F\right), \\
%     &= \lambda^2\left(\sum_{i\neq k}\langle\bm{T}_i^k,\bm{T}_i^k\rangle_F + \sum_{i,j\neq k}\langle\bm{T}_i,\bm{T}_j^k\rangle_F\right),\\
%     &= \lambda^2\left(\sum_{i\neq k}\|\bm{T}_i^k\|_F^2+2\sum_{\substack{1 \leq i < j \leq n \\ i, j \neq k}}\langle\bm{T}_i^k,\bm{T}_j^k\rangle_F \right),\\
%     &= \lambda^2\left(\sum_{i\neq k}\|\bm{T}_i^k\|_F^2+2\sum_{\substack{1 \leq i < j \leq n \\ i, j \neq k}}\|\bm{T}_i^k\|_F\|\bm{T}_j^k\|_F\cos(\bm{T}_i^k,\bm{T}_j^k)\right).
% \end{align}
\section{Additional Analysis} \label{sec:extra_analysis}
\subsection{GPT-2 Text Classification Experiments}
We evaluate our proposed methods against established baselines by compressing seven independently trained GPT-2 models on text classification tasks. As shown in Table~\ref{gpt2-table}, both our \textit{RanDeS} algorithms achieved significantly higher classification accuracy while doubling the memory footprint, consistent with the high performance on other benchmarks.
\begin{table*}[h]
    % \caption{Performance comparison on seven GLUE text classification tasks with GPT-2 models merging and compression, \rbt{along with} memory footprint estimate\rbt{s}. 
    % \rbt{In the \textit{Avg.(\%)} column, the values in parentheses indicate performance relative to the fine-tuned model (set to 100\%). In the \textit{Bits(Gb)} column, the values in parentheses represent the relative model size compared to the pre-trained model (set to 1.00).}
    % Three repetitions are done for methods with randomness\rbt{; v}ariances smaller than $0.1\%$ are omitted.}
    \caption{\rbt{Performance and memory comparison of GPT-2 models across seven GLUE text classification tasks, showing absolute and normalized accuracy (\%), as well as memory footprint (Gb). Results averaged over three runs where applicable. Variances smaller than 0.1\% are omitted.}}
    \label{gpt2-table}
    \begin{center}
        \resizebox{\linewidth}{!}{
            \begin{tabular}{l|cc|*{7}{c}}
                \toprule
                \multicolumn{1}{c}{Method} & \multicolumn{1}{c}{Avg.(\%) \(\uparrow\)} & \multicolumn{1}{c}{Bits(Gb) $\downarrow$} & \multicolumn{1}{c}{CoLA} & \multicolumn{1}{c}{MNLI} & \multicolumn{1}{c}{MRPC} & \multicolumn{1}{c}{QNLI} & \multicolumn{1}{c}{QQP} & \multicolumn{1}{c}{RTE} & \multicolumn{1}{c}{SST-2} \\
                \midrule
\rowcolor{gray!15}Pre-trained  & 44.5 \scriptsize{(54.3)} & 0.498 \scriptsize{(1.00)} & 30.9 & 33.0 & 31.4 & 49.2 & 63.2 & 52.7 & 50.9 \\
Weight Averaging      & 56.1 \scriptsize{(63.3)} & 0.498 \scriptsize{(1.00)} & 55.0 & 55.1 & 51.0 & 57.6 & 76.7 & 44.8 & 52.5 \\
Fisher Merging     & 58.7 \scriptsize{(64.7)} & 0.498 \scriptsize{(1.00)} & 54.8 & 58.0 & 39.5 & 63.3 & 81.5 & 49.1 & 64.7 \\
RegMean           & 68.8 \scriptsize{(79.7)} & 0.498 \scriptsize{(1.00)} & 61.7 & 70.4 & 65.4 & 69.7 & 78.8 & 56.0 & 79.7 \\
Task Arithmetic     & 70.0 \scriptsize{(85.4)} & 0.498 \scriptsize{(1.00)} & 68.7 & 68.6 & \underline{69.6} & 70.5 & 81.8 & 47.3 & 83.6 \\
Ties-Merging      & 70.0 \scriptsize{(82.4)} & 0.498 \scriptsize{(1.00)} & 68.4 & 71.4 & 68.4 & 69.6 & 82.4 & 47.7 & 81.8 \\
\midrule
\rowcolor{gray!15}Fine-tuned   & 82.0 \scriptsize{(100)} & 3.49 \scriptsize{(7.00)} & 76.8 & 82.1 & 80.4 & 88.3 & 89.6 & 65.3 & 91.2 \\
\textbf{RanDeS-S (Ours)}   & \textbf{76.7} \scriptsize{\textbf{(93.5)}} &  0.997 \scriptsize{(2.00)} & \textbf{71.6} & \underline{80.3} & \textbf{73.9} & \underline{85.8} & \underline{88.5} & 47.5 & 89.3 \\
\textbf{RanDeS-RSF (Ours)}  & $71.3 \pm 0.6$ \text{ \scriptsize{(87.0)}} & 0.997 \scriptsize{(2.00)} & $62.3 \pm 0.3$ & 78.2 & $46.1 \pm 4$ & 82.6 & 88.4 & \underline{52.7} & \underline{88.9} \\
\textbf{RanDeS-SRSF (Ours)}   & $\underline{76.6} \pm \underline{0.2}$ \text{ \scriptsize(\underline{93.4})} &  0.997 \scriptsize{(2.00)} & $\underline{70.3} \pm \underline{0.1}$ & \textbf{81.0} & $61.0 \pm 1.3$ & \textbf{87.2} & \textbf{89.3} & $\textbf{57.5} \pm \textbf{0.3}$ & \textbf{90.2} \\
                \bottomrule
            \end{tabular}
        }
    \end{center}
\end{table*}

\subsection{Dynamics between Layer Shuffling and Random Sign Flips}
We investigate how varying layer shuffling and random sign flips parameters affects model performance. We test target layer skip rates of ${1, 2, 3, 4}$, where only every k-th target layer within repetitive layer sets is shuffled and sign-flipped. We also introduce \textit{layer shifting} -- a deterministic alternative to shuffling that shifts layers one position deeper with wrap-around -- to study how different decorrelation approaches affect performance.

\begin{figure}[t]
    \centering
    \includegraphics[width=1.0\textwidth]{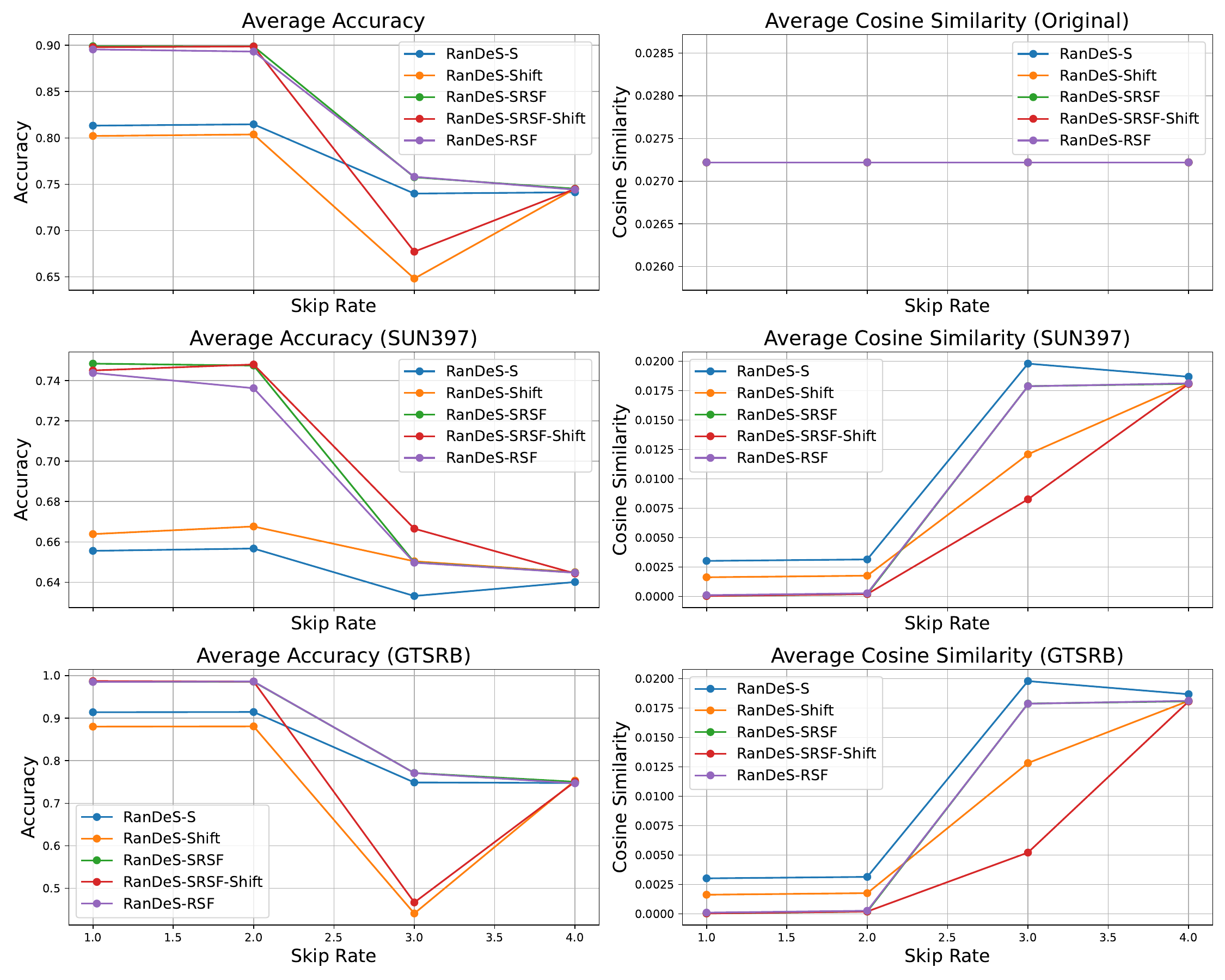}
    \vspace{-4mm}
    \caption{Average accuracy and cosine similarity among interfering deltas when retrieving SUN397 and GTSRB models from 8 merged CLIP-ViT-B/32 models with various target layer skipping rates and shuffling/superposition setups.}
\label{fig:shuffle_superpose_levels}
    \vspace{-1mm}
\end{figure}

Experiments on eight CLIP-ViT-B/32 benchmarks (Figure~\ref{fig:shuffle_superpose_levels}) show averaged results across three repetitions, focusing on overall benchmark accuracy and two specific tasks: SUN397~\citep{sun397} and GTSRB~\citep{stallkamp2012man}. We analyze both task performance and average pairwise cosine similarity among deltas - both original and among interfering deltas during model retrieval.

As skip rate increases, accuracy declines while cosine similarity rises. Performance remains stable up to skip rate 2, suggesting potential memory savings through selective layer manipulation.
For GTSRB, \textit{RanDeS-S} outperforms \textit{RanDeS-Shift} despite higher cosine similarity, indicating the method of achieving orthogonality matters beyond decorrelation levels.
This pattern reverses for SUN397, revealing task-dependent variations and opportunities for task-specific optimization.

\begin{figure}[h]
    \centering
    \includegraphics[width=1.0\textwidth]{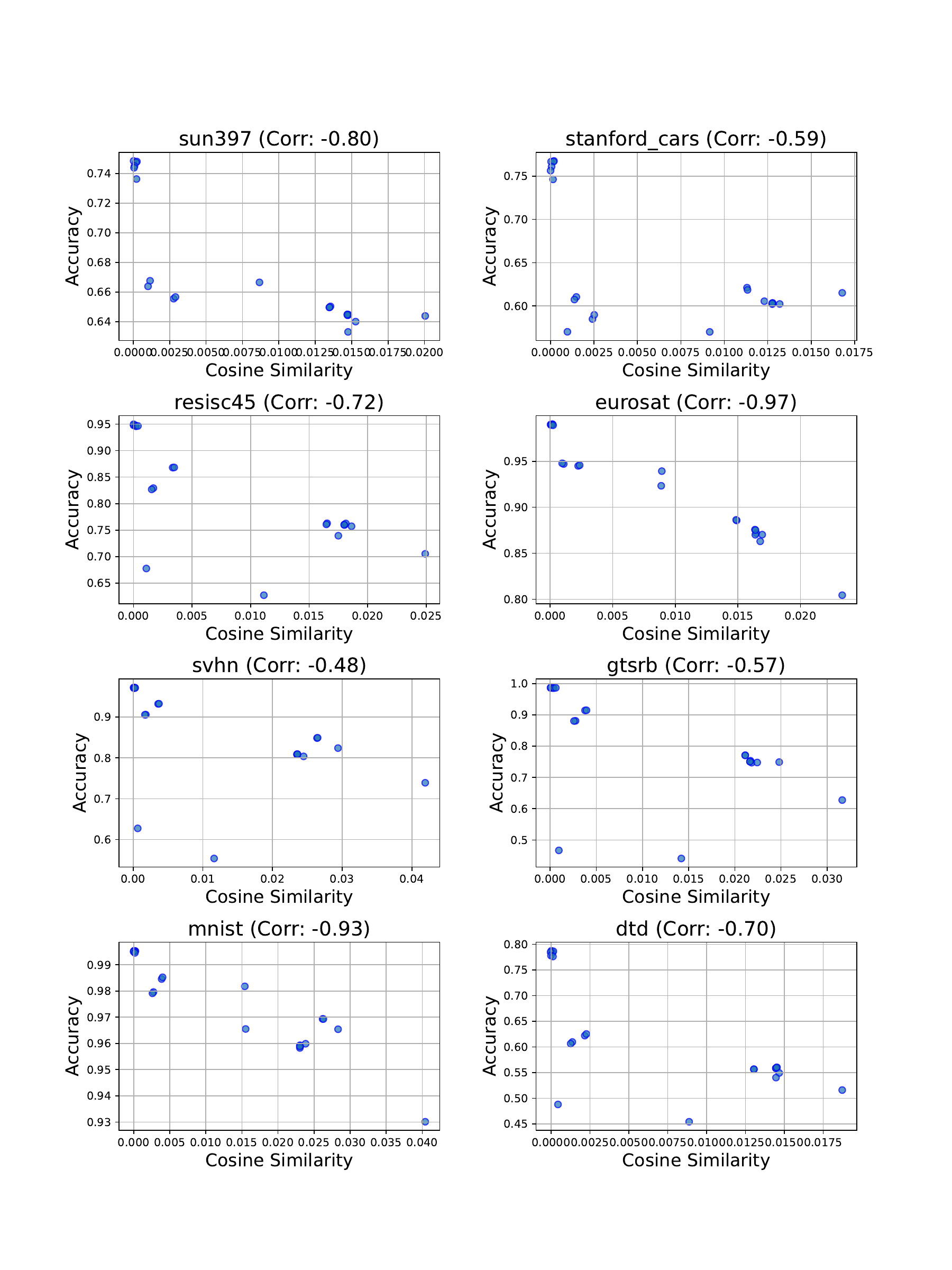}
    \vspace{-16mm}
    \caption{Correlation between the pairwise cosine similarity among interfering deltas and the accuracy on the eight image classification tasks, with CLIP-ViT-B/32 merged with different levels of shuffling and superposition.}
\label{fig:cos_sim_acc_correlation}
    \vspace{-1mm}
\end{figure}

The correlation between interfering deltas' cosine similarity and accuracy shows a negative trend (Figure~\ref{fig:cos_sim_acc_correlation}), most pronounced in EuroSAT~\citep{helber2019eurosat} and MNIST~\citep{deng2012mnist}. While patterns vary across tasks, peak accuracy consistently occurs near zero cosine similarity. This observation, combined with \textit{RanDeS-SRSF}'s strong performance at low skip rates (Figure~\ref{fig:shuffle_superpose_levels}), suggests a cosine similarity threshold may exist above which method selection and task properties become less critical.

\end{document}